\documentclass[electronics,article,accept,pdftex,moreauthors]{Definitions/mdpi}

\firstpage{1} 
\pubvolume{1}
\issuenum{1}
\articlenumber{0}
\pubyear{2025}
\copyrightyear{2025}
\externaleditor{Domenico Rosaci} % More than 1 editor, please add `` and '' before the last editor name
\datereceived{7 January 2025} 
\daterevised{2 February 2025} % Comment out if no revised date
\dateaccepted{4 February 2025} 
\datepublished{ } 
\usepackage[ruled,linesnumbered]{algorithm2e}
\usepackage{algpseudocode}
\usepackage{xcolor}
\hreflink{https://doi.org/} % If needed use \linebreak
\Title{A Novel Trustworthy Video Summarization Algorithm Through a Mixture of LoRA Experts} %EE: Attn: title altered

\TitleCitation{A Novel Trustworthy Video Summarization Algorithm a Through Mixture of LoRA Experts}
  
\AuthorNames{Wenzhuo Du and Gerun Wang}

\Author{{Wenzhuo Du} %MDPI: Please carefully check the accuracy of names and affiliations. Please notice that any changes about authorship, affiliations, and corresponding authors are not allowed after the paper has been accepted, including adding, removing, or changing the order.
 $^{1}$, Gerun Wang $^{1}$, Xin Li $^{1,2,}$*, Guancheng Chen $^{3}$, Jian Gao $^{1,2}$ and Hang Zhao $^{3}$}

\address{%
$^{1}$ \quad School of Information and Network Security, People’s Public Security University of China, Beijing 102206, China; \\
$^{2}$ \quad Key Laboratory of Security Prevention Technology and Risk Assessment of the Ministry of Public Security, Beijing 100038, China\\
$^{3}$ \quad School of Criminal Investigation, People’s Public Security University of China, Beijing 102206, China; 
}

\corres{Correspondence: lixin@ppsuc.edu.cn}

\abstract{With the exponential growth of user-generated content on video-sharing platforms, the challenge of facilitating efficient searching and browsing of videos has garnered significant attention. To enhance users' ability to swiftly locate and review pertinent videos, the creation of concise and informative video summaries has become increasingly important. Video-llama is an effective tool for generating video summarization, but it cannot effectively unify and optimize the modeling of temporal and spatial features and requires a lot of computational resources and time. Therefore, we propose MiLoRA-ViSum to more efficiently capture complex temporal dynamics and spatial relationships inherent in video data and to control the number of parameters for training.   By extending traditional Low-Rank Adaptation (LoRA) into a sophisticated mixture-of-experts paradigm, MiLoRA-ViSum incorporates a dual temporal--spatial adaptation mechanism tailored specifically for video summarization tasks. This approach dynamically integrates specialized LoRA experts, each fine-tuned to address distinct temporal or spatial dimensions.   Extensive evaluations of the VideoXum and ActivityNet datasets demonstrate that MiLoRA-ViSum achieves the best summarization performance compared to state-of-the-art models, while maintaining significantly lower computational costs. The proposed mixture-of-experts strategy, combined with the dual adaptation mechanism, highlights the model’s potential to enhance video summarization capabilities, particularly in large-scale applications requiring both efficiency and precision.}

\keyword{\textls[-15]{mixture of LoRA experts;%MDPI: Please check if the uppercase of the this keyword should be kept?
 video summarization; low-rank adaptation; temporal--spatial adaptation;} Video-LLaMA; computational efficiency; transformer models} 

\usepackage{fancyhdr}
\fancypagestyle{plain}{%
    \fancyhf{} % 清空当前的页眉和页脚设置
}

\fancypagestyle{mainstyle}{%
    \fancyhf{} % 清空当前的页眉和页脚设置
}
\begin{document}

%%%%%%%%%%%%%%%%%%%%%%%%%%%%%%%%%%%%%%%%%%

\section{Introduction}

Video summarization, which involves distilling lengthy video content into concise and informative summaries, has become increasingly significant due to the exponential growth of video data across various domains. Applications such as content browsing, video analytics, and~archival management require efficient summarization techniques, as~manually reviewing entire video datasets is often unfeasible~\cite{saini2023video, he2023align, jangra2023survey}. Traditional summarization methods, predominantly relying on heuristic strategies like clustering or keyframe extraction, are computationally efficient but lack the ability to generalize effectively across diverse video content. These approaches often fail to capture the complex, context-dependent information embedded within videos, as they primarily focus on either temporal or spatial features in isolation rather than addressing the inherent interplay between the two~\cite{elfeki2022multi}. This limitation results in summaries that are often incomplete or fail to convey the essential semantics of the original content. The increasing complexity of video datasets necessitates models that can process and unify temporal and spatial features in a computationally efficient manner, a~challenge that remains inadequately addressed by existing approaches.

Recent advancements in deep learning have greatly enhanced video summarization capabilities, with transformer-based models emerging as a cornerstone of these developments. Among such models, Video-LLaMA has shown significant promise due to its ability to effectively capture long-range dependencies and complex temporal dynamics embedded within video sequences. This capacity enables it to process extensive video content while maintaining semantic coherence and temporal consistency. However, despite its notable performance, the~high computational costs and extensive resource requirements for training and fine-tuning these large-scale models have posed significant barriers to their practical deployment. Addressing these limitations necessitates an innovative approach that balances the need for high performance with computational efficiency. In~this context, we propose transitioning from traditional Low-Rank Adaptation (LoRA) to a Mixture of LoRA Experts (MiLoRA) framework, which distributes computational workloads across specialized low-rank adaptation modules. This approach not only retains the advantages of Video-LLaMA but also introduces a scalable and resource-efficient solution for real-world applications~\cite{zhang2023video}.

To address these challenges, we propose a novel approach to video summarization, named MiLoRA-ViSum, which leverages a Mixture of Low-Rank Adaptation (MiLoRA) modules integrated into the Video-LLaMA backbone. MiLoRA extends the conventional LoRA framework by employing multiple specialized low-rank adaptation experts, each tailored to capture distinct aspects of temporal and spatial features in video data. These experts dynamically collaborate during training and inference, enabling the model to efficiently adapt to the complex dependencies inherent in video sequences. By~distributing the adaptation workload across multiple experts, MiLoRA-ViSum not only enhances the model's adaptability but also significantly reduces the number of trainable parameters while maintaining competitive performance. This approach ensures the generation of high-quality video summaries with minimized computational overhead, addressing the practical constraints of large-scale video summarization~tasks.

In this paper, we evaluate the performance of MiLoRA-ViSum, a~model that integrates the Mixture of Low-Rank Adaptation (MiLoRA) framework, using two widely recognized and diverse video summarization datasets, VideoXum~\cite{lin2023videoxum} and ActivityNet~\cite{caba2015activitynet}. These datasets were chosen for their ability to present a range of challenging scenarios, encompassing varied video content and temporal--spatial complexities, thereby providing a robust benchmark for assessing the proposed method's adaptability and effectiveness. Through extensive experimentation, MiLoRA-ViSum demonstrates its capability to generate high-quality video summaries while achieving computational efficiency. Notably, the~results reveal that MiLoRA-ViSum not only matches but, in~several cases, surpasses the performance of state-of-the-art models in terms of summarization quality. This is accomplished with a significantly reduced computational footprint, underscoring the method’s potential for scalable and efficient deployment in real-world applications. These findings highlight the practical advantages of incorporating multiple specialized low-rank adaptation experts to optimize temporal and spatial feature learning, validating MiLoRA-ViSum as a promising advancement in video summarization~research.

To address these challenges, we propose a novel approach named MiLoRA-ViSum, which extends the concept of Low-Rank Adaptation (LoRA) by introducing a mixture of LoRA experts framework with a dual adaptation mechanism specifically tailored for video summarization tasks. Unlike traditional LoRA implementations that apply a singular low-rank adaptation, MiLoRA-ViSum incorporates multiple specialized LoRA experts, each fine-tuned for distinct aspects of temporal and spatial feature processing. This multi-expert system dynamically selects and integrates the most relevant adaptations based on the video content, enabling the model to better capture the complex dependencies and interactions inherent in video data. The dual-adaptation mechanism further enhances the framework by applying specialized LoRA adaptations to both the temporal attention layers and spatial convolutional layers of the Video-LLaMA backbone. This comprehensive approach allows the model to effectively process both the sequential patterns and spatial structures present in video sequences, achieving a holistic and computationally efficient representation. To~the best of our knowledge, MiLoRA-ViSum is the first work to integrate a mixture-of-experts strategy within the LoRA framework for simultaneous temporal and spatial fine-tuning, marking a significant methodological advancement in video summarization~research.

 \vspace{3pt}
\textbf{{In summary, our key contributions are as follows:}}%MDPI: Please confirm if the bold is unnecessary and can be removed. The following highlights are the same.

\begin{itemize}
    \item We propose \textbf{{MiLoRA-ViSum}}, a~novel framework that introduces a mixture of LoRA experts to achieve an advanced dual temporal--spatial adaptation mechanism tailored specifically for video summarization tasks. By~integrating specialized LoRA experts for temporal and spatial layers, our approach ensures efficient and precise feature capture across video data.
    \item We demonstrate significant methodological advancements by leveraging the mixture of LoRA experts to dynamically optimize both temporal attention layers and spatial convolutional layers within the Video-LLaMA backbone. This dual adaptation mechanism enhances the ability to model complex temporal--spatial dependencies while maintaining computational efficiency.
    \item We provide a comprehensive empirical evaluation of MiLoRA-ViSum on two widely recognized video summarization datasets, VideoXum and ActivityNet, conducting a thorough comparison with existing state-of-the-art models. Our results indicate that MiLoRA-ViSum achieves competitive performance while reducing the number of trainable parameters significantly, underscoring its scalability and practicality for real-world applications.
\end{itemize}

\section{Related~Work}
\unskip
\subsection{Video~Summarization}

Video summarization has become a critical research area within computer vision, driven by the exponential increase in video content across various platforms and domains. This task focuses on extracting concise, representative summaries from extensive video sequences, enabling efficient content consumption for applications such as video analytics, content recommendation systems, and~archival retrieval~\cite{rennard2023abstractive}. Traditional video summarization methods often rely on heuristic approaches, including clustering techniques and keyframe extraction algorithms. These methods aim to identify visually distinct or statistically significant frames within a video~\cite{potapov2014category, yang2024ostr}. While such approaches are computationally efficient and have seen widespread adoption in earlier systems, their reliance on simple rules and thresholds frequently limits their ability to capture complex patterns in real-world video data. These methods often fail to generalize across diverse video types and typically overlook the contextual information and temporal relationships embedded within videos, resulting in summaries that are disjointed or incomplete~\cite{selva2023video, yang2021condensenet}.

The advent of deep learning has revolutionized video summarization by introducing methods capable of modeling spatio-temporal dependencies in a more nuanced and data-driven manner. Early attempts in this space employed Convolutional Neural Networks (CNNs) to capture spatial features and Recurrent Neural Networks (RNNs) to model temporal dynamics. These architectures demonstrated improved performance compared to heuristic-based methods by leveraging large-scale datasets to learn meaningful feature representations. For~example, RNNs, such as Long Short-Term Memory (LSTM) networks~\cite{zhang2016video, zheng2023dynamic}, have been effective in capturing temporal sequences, enabling a better understanding of video narratives and transitions~\cite{apostolidis2021video}. However, these methods are often constrained by their limited ability to process long video sequences due to vanishing gradient problems and high computational costs associated with recurrent architectures. As~a result, while CNN-RNN-based methods represent a step forward, they still fall short in addressing the full spectrum of challenges in video summarization, particularly for complex, dynamic video content~\cite{ma2002user}.

More recently, transformer-based models, known for their self-attention mechanisms, have emerged as a promising solution to the limitations of earlier methods. These models excel at capturing long-range dependencies and hierarchical relationships in sequential data, making them well-suited for video summarization tasks~\cite{apostolidis2021video}. Models such as Video-LLaMA leverage transformers to process both temporal and spatial features in a unified manner, enabling them to capture intricate dependencies across frames while maintaining contextual coherence. However, despite their impressive performance improvements, transformer-based models are associated with high computational costs and substantial resource requirements during training and inference~\cite{haq2020video}. These constraints hinder their deployment in resource-limited environments and large-scale real-world applications. Addressing these challenges requires innovative strategies that balance computational efficiency with the ability to capture rich spatio-temporal relationships~\cite{chu2015video}. Consequently, recent research efforts have focused on developing more efficient transformer architectures, parameter reduction techniques, and~hybrid models to make video summarization more accessible and scalable without compromising performance~\cite{saini2023video}.

\subsection{LoRA in Vision-Language~Models}

Low-rank adaptation (LoRA) has emerged as a key technique in vision--language models, providing a mechanism to significantly reduce computational overhead while retaining model performance. Originally introduced for efficiently fine-tuning language models, LoRA works by introducing low-rank updates to the weight matrices of pre-trained models, which allows for task-specific adaptation without retraining the entire network. This approach has proven particularly beneficial in the context of vision-language models, where the computational complexity is often exacerbated by the need to process both visual and textual inputs. For~instance, in VideoLLM-online, LoRA was successfully employed to adapt pre-trained large language models for streaming video understanding tasks. This integration allowed for real-time processing of video streams~\cite{zanella2024low}, reducing the memory and computational requirements associated with traditional fine-tuning methods. By~focusing on low-rank updates, the~approach enables the model to generalize effectively across tasks while maintaining high accuracy and efficiency~\cite{lu2024adaptive}.

Recent advancements in LoRA-based techniques have further demonstrated their versatility in vision--language tasks, particularly in models such as Video-LLaMA, where LoRA facilitates the integration of visual and textual modalities. In~Video-LLaMA, LoRA enables the fine-tuning of pre-trained language models to incorporate visual information, bridging the gap between the two modalities~\cite{gou2023mixture}. This model excels in tasks such as video question answering and caption generation, where understanding both spatial and temporal visual features is critical. By~leveraging LoRA, these models manage to adapt pre-trained architectures to complex video-based tasks with minimal additional computational cost. However, despite their success, these applications primarily focus on tasks involving generalized video understanding rather than the more specialized task of video summarization. This highlights a gap in the literature where LoRA's potential for optimizing both temporal and spatial adaptations in the context of summarization remains largely untapped~\cite{elgendy2024geollava}.

Our work seeks to address this gap by extending the application of LoRA to video summarization tasks, with~a specific focus on optimizing both temporal and spatial feature representations. While previous studies have demonstrated the utility of LoRA in simplifying model adaptation for vision-language tasks~\cite{jiang2024vlm2vec}, they often overlook the unique challenges posed by video summarization. For~instance, summarization requires not only understanding the visual content in individual frames but also capturing the temporal progression and contextual transition throughout a video. To~address this, we introduce a novel approach that incorporates a Mixture of LoRA Experts (MiLoRA). Unlike traditional LoRA methods that apply uniform low-rank adaptations across layers, MiLoRA employs specialized adaptations for different components of the video summarization pipeline. This includes tailoring LoRA modules to temporal attention mechanisms for modeling long-range dependencies and spatial convolutional layers for capturing localized visual details. By leveraging this mixture of experts, our method provides a more granular and effective adaptation strategy, enabling the generation of concise and coherent summaries while maintaining computational efficiency~\cite{chen2024benchmarking}. This approach represents a significant advancement in the application of LoRA to specialized video tasks, paving the way for broader adoption in computationally constrained~environments.

\section{Method}
\unskip

\subsection{MiLoRA-ViSum Model~Architecture}

\begin{figure}[H]
%\centering
\includegraphics[width=1\textwidth]{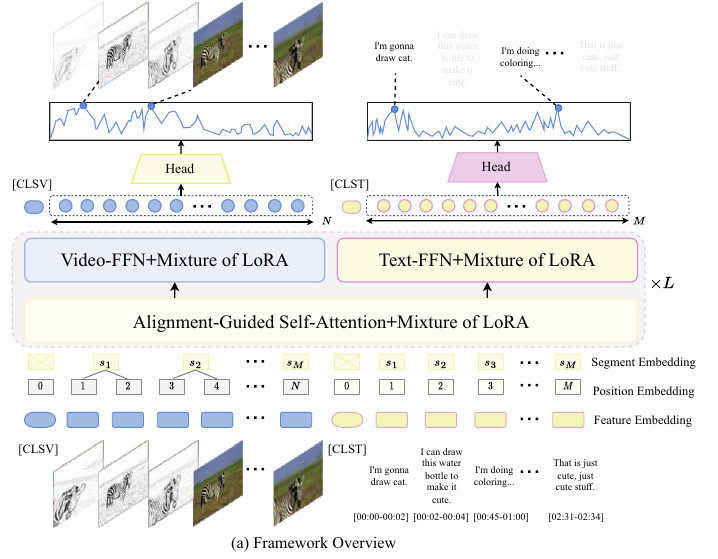}
\caption{{MiLoRA-ViSum} %MDPI: Figure 1 is moved under where it's first mentioned in the paper. Please confirm.
%MDPI: Please remove the subfigure label (a) in the image. 
 model architecture.}
\label{fig:model_architecture}
\end{figure}

The proposed \textbf{{MiLoRA-ViSum}} framework extends the Low-Rank Adaptation (LoRA) methodology by integrating a Mixture of LoRA Experts (MiLoRA) into the Video-LLaMA backbone model, significantly enhancing the model's efficiency and performance in video summarization tasks (see Figure~\ref{fig:model_architecture}). Unlike the traditional LoRA approach, which applies uniform low-rank updates to model layers, MiLoRA introduces a mixture of specialized low-rank modules tailored for different components of the architecture. This ensures that both temporal and spatial features of video data are effectively captured while maintaining computational efficiency. In this paper, we use the MiLoRA-ViSum framework to generate video summaries as shown in the pseudo-code in Algorithm \ref{alg:myalgorithm}.

The MiLoRA mechanism operates by introducing multiple expert modules, each designed to adaptively fine-tune specific layers of the Video-LLaMA model. For~a given weight matrix \( \mathbf{W}_\ell \) in the \( \ell \)-th layer, the~adaptation is expressed as a summation of low-rank updates contributed by \( K \) specialized experts:
\begin{equation}
\Delta \mathbf{W}_\ell = \sum_{k=1}^K \mathbf{B}_\ell^{(k)} \mathbf{A}_\ell^{(k)},
\end{equation}
where \( \mathbf{B}_\ell^{(k)} \in \mathbb{R}^{d \times r_k} \) and \( \mathbf{A}_\ell^{(k)} \in \mathbb{R}^{r_k \times k} \) are the low-rank decomposition matrices for the \( k \)-th expert, with~\( r_k \) representing the rank for that expert. The~summation allows MiLoRA to flexibly aggregate updates from multiple experts, each of which specializes in distinct aspects of the video summarization~process.

\begin{algorithm}[H]
	%\rightskip=-1em
	\caption{MiLoRA-ViSum}
        \label{alg:myalgorithm}
	\KwIn{~\\~~~$ \bullet $ Video Frames $ V  $ \\
		~~~$ \bullet $ Pre-trained Video-LLaMA Model  $ M_{\rm{base}} $\\
	    ~~~$ \bullet $ MiLoRA Configuration $ C $  }
	\KwOut{~\\~~~$ \bullet $ Summarized Video Content $ S $}
	Initialize Mixture of LoRA Experts $ \left\{ {{\rm{B}^{\left( k \right)}},{\rm{A}^{\left( k \right)}}} \right\}_{k = 1}^K $ \\
	Initialize Temporal Attention and Spatial Convolutional Layers $ \left\{ {{{\rm{W}}_{{\rm{temporal}}}},{{\rm{W}}_{{\rm{spatial}}}}} \right\} $ \\
	Load pre-trained weights from $ M_{\rm{base}} $ \\
	\For{each Video Sequence $ v \in V $}{a. \textbf{Temporal Feature Extraction:} \\
		~~~Compute temporal attention outputs using:\\
		~~~~$ {{\rm{y}}_{{\rm{temporal}}}} = {\rm{Soft}}\max \left( {\frac{{{\rm{Q}}{{\rm{K}}^{\rm{T}}}}}{{\sqrt {{d_k}} }} + \Delta {{\rm{W}}_{{\rm{temporal}}}}} \right){\rm{V}} $
		
		b. \textbf{Spatial Feature Extraction:}\\
		~~~Apply spatial convolution with low-rank updates:\\
		~~~~$ {{\rm{y}}_{{\rm{spatial}}}} = \sigma \left( {{\rm{X}}*\left( {{{\rm{W}}_{{\rm{spatial}}}} + \Delta {{\rm{W}}_{{\rm{spatial}}}}} \right)} \right) $
		
		c. \textbf{Fusion of Features:}\\
		~~~Fuse temporal and spatial outputs:\\
		~~~~$ {{\rm{y}}_{{\rm{fused}}}} = \alpha  \cdot {{\rm{y}}_{{\rm{temporal}}}} + \left( {1 - \alpha } \right) \cdot {{\rm{y}}_{{\rm{spatial}}}} $
		
		d. Generate summary representation $ \rm{s} $ using $ M_{\rm{base}} $ with fused features.
		
		e. Append $ \rm{s} $ to $ S $. 
	}
	
	\textbf{return} $ S $
\end{algorithm}
\vspace{12pt}

To capture the distinct characteristics of {visual and textual data}, MiLoRA-ViSum employs separate modules for video and text processing. These modules are integrated into the Video-Feedforward Networks (Video-FFN) and Text-Feedforward Networks \mbox{(Text-FFN),} respectively, as~shown in Figure~\ref{fig:model_architecture}. For~the video stream, the adaptation focuses on capturing temporal dynamics through self-attention mechanisms augmented with low-rank updates. For~a video input sequence \( \mathbf{x}_v \), the~output is computed as:
\begin{equation}
\mathbf{y}_v = \text{Softmax}\left(\frac{\mathbf{Q}_v \mathbf{K}_v^\top}{\sqrt{d_k}} + \Delta \mathbf{W}_{v,\text{attn}}\right)\mathbf{V}_v,
\end{equation}
where \( \mathbf{Q}_v, \mathbf{K}_v, \mathbf{V}_v \) are the query, key, and~value matrices for the video input, and~\( \Delta \mathbf{W}_{v,\text{attn}} \) represents the low-rank updates applied specifically to the attention~mechanism.

Similarly, for~textual inputs \( \mathbf{x}_t \), the~Text-FFN modules leverage LoRA to fine-tune pre-trained language model layers, enabling alignment between visual and textual modalities. The~adaptation for the text attention layers is given by:
\begin{equation}
\mathbf{y}_t = \text{Softmax}\left(\frac{\mathbf{Q}_t \mathbf{K}_t^\top}{\sqrt{d_k}} + \Delta \mathbf{W}_{t,\text{attn}}\right)\mathbf{V}_t.
\end{equation}

A key innovation in the MiLoRA-ViSum architecture is the alignment-guided self-attention mechanism~\cite{laurenccon2024matters}, which bridges the gap between video and text streams. This module integrates temporal and textual embeddings by leveraging segment embeddings, position embeddings, and~feature embeddings, as~illustrated in Figure~\ref{fig:model_architecture}. The~alignment is achieved by computing attention scores that emphasize cross-modal dependencies:
\begin{equation}
\text{Attention}(\mathbf{X}_v, \mathbf{X}_t) = \text{Softmax}\left(\frac{\mathbf{Q} (\mathbf{K}_v + \mathbf{K}_t)^\top}{\sqrt{d_k}}\right),
\end{equation}
where \( \mathbf{Q} \) represents the combined query matrix derived from both modalities, and~\( \mathbf{K}_v, \mathbf{K}_t \) are the key matrices for video and text, respectively. The~alignment ensures that the final embeddings reflect meaningful interactions between the two~streams.

To train the MiLoRA-ViSum framework effectively, we employ a composite loss function that combines a summarization loss \( \mathcal{L}_{\text{sum}} \) with an expert regularization term \( \mathcal{L}_{\text{reg}} \). The~overall loss function is defined as:
\begin{equation}
\mathcal{L} = \mathcal{L}_{\text{sum}} + \lambda_{\text{reg}} \sum_{\ell=1}^L \sum_{k=1}^K \left(\|\mathbf{B}_\ell^{(k)}\|_F^2 + \|\mathbf{A}_\ell^{(k)}\|_F^2\right),
\end{equation}
where \( \lambda_{\text{reg}} \) controls the contribution of the regularization term, ensuring that the low-rank updates remain compact and interpretable. This regularization encourages sparsity among the expert modules, enabling efficient utilization of computational~resources.

Figure~\ref{fig:model_architecture} provides a comprehensive overview of the MiLoRA-ViSum framework. The~figure highlights the modular structure of the architecture, with~separate branches for video and text processing, and~illustrates how alignment-guided self-attention integrates these modalities~\cite{wang2024visionllm}. The~inclusion of segment embeddings and positional information ensures that both temporal and spatial dependencies are preserved. Additionally, the~visual representation of expert modules emphasizes their modularity and adaptability, which are central to the efficiency of the MiLoRA~approach.

\subsection{Integrated Temporal--Spatial Adaptation and Optimization for Video~Summarization}

{
A significant challenge in video summarization is the effective integration of temporal and spatial features within video sequences, as~these dimensions exhibit complex interdependencies. To~address this, we propose a novel dual adaptation mechanism that extends the MiLoRA (Mixture of LoRA Experts) framework to both temporal attention layers and spatial convolutional layers of the Video-LLaMA model. This design allows for the simultaneous fine-tuning of temporal and spatial features, capturing intricate dependencies while maintaining computational efficiency.
For the temporal attention layer \( \ell_t \), we model the low-rank adaptation as:}
\begin{equation}
	\Delta \mathbf{W}_{\ell_t} = \sum_{k=1}^K \mathbf{B}_{\ell}^{(k)} \mathbf{A}_{\ell}^{(k)},
\end{equation}
{where \( \mathbf{B}_{\ell}^{(k)} \in \mathbb{R}^{d \times r_k} \) and \( \mathbf{A}_{\ell}^{(k)} \in \mathbb{R}^{r_k \times d} \) represent the decomposition matrices for the \( k \)-th expert, and~\( K \) denotes the total number of experts. Each expert specializes in capturing a specific aspect of temporal dynamics, allowing for a flexible aggregation of information across different temporal patterns. The~output of the temporal attention mechanism, incorporating the MiLoRA updates, is computed as:}
\begin{equation}
	\mathbf{y}_{\ell_t} = \text{Softmax}\left(\frac{\mathbf{Q}_{\ell} \mathbf{K}_{\ell}^\top}{\sqrt{d_k}} + \Delta \mathbf{W}_{\ell_t}\right)\mathbf{V}_{\ell},
\end{equation}
{where \( \mathbf{Q}_{\ell}, \mathbf{K}_{\ell}, \mathbf{V}_{\ell} \) represent the query, key, and~value matrices for the temporal attention layer, respectively.
Similarly, for~the spatial convolutional layer \( \ell_s \), we apply a low-rank adaptation using the same formulation:}
\begin{equation}
	\Delta \mathbf{W}_{\ell_s} = \sum_{k=1}^K \mathbf{B}_{\ell}^{(k)} \mathbf{A}_{\ell}^{(k)},
\end{equation}
{where \( \mathbf{B}_{\ell}^{(k)} \in \mathbb{R}^{d \times r_k} \) and \( \mathbf{A}_{\ell}^{(k)} \in \mathbb{R}^{r_k \times d} \) are the low-rank matrices for the \( k \)-th expert in the spatial layer. This decomposition enables the efficient modeling of spatial dependencies without significantly increasing computational costs. The~spatial feature transformation is expressed as:}
\begin{equation}
	\mathbf{y}_{\ell_s} = \sigma\left(\mathbf{X}_{\ell} \ast \left(\mathbf{W}_{\ell} + \Delta \mathbf{W}_{\ell_s}\right)\right),
\end{equation}
{where \( \mathbf{X}_{\ell} \) is the spatial input feature map, \( \ast \) represents the convolution operation, and~\( \sigma(\cdot) \) is the activation function.
A key contribution of our dual adaptation mechanism is the joint optimization of temporal and spatial adaptations. Instead of treating these adaptations independently, we introduce a unified loss function that balances their contributions, enabling cohesive learning across both dimensions. The~overall loss function for the MiLoRA framework is now comprehensively described as:}
\begin{equation}
	\mathcal{L}(\Theta) = \mathcal{L}_{\text{sum}} + \lambda_t \sum_{\ell_t} \sum_{k=1}^K \left(\|\mathbf{B}_{\ell}^{(k)}\|_F^2 + \|\mathbf{A}_{\ell}^{(k)}\|_F^2\right) + \lambda_s \sum_{\ell_s} \sum_{k=1}^K \left(\|\mathbf{B}_{\ell}^{(k)}\|_F^2 + \|\mathbf{A}_{\ell}^{(k)}\|_F^2\right),
\end{equation}
{where \( \mathcal{L}_{\text{sum}} \) represents the summarization loss, \( \lambda_t \) and \( \lambda_s \) are regularization coefficients for the temporal and spatial layers, respectively, and~\( \|\cdot\|_F \) denotes the Frobenius norm. This unified formulation, aligned with the explanation in Section~\ref{sec3.3}, ensures that the adaptations are both effective and computationally efficient while promoting sparsity and interpretability among expert modules.}

The dual adaptation mechanism leverages the interdependencies between temporal and spatial features by aligning their respective representations. Temporal attention outputs \( \mathbf{y}_{\ell_t} \) are fused with spatial convolutional outputs \( \mathbf{y}_{\ell_s} \) to create a unified representation:
\begin{equation}
	\mathbf{y}_{\text{fused}} = \alpha \cdot \mathbf{y}_{\ell_t} + (1 - \alpha) \cdot \mathbf{y}_{\ell_s},
\end{equation}
where \( \alpha \) is a learnable weight parameter balancing the contributions of temporal and spatial features. This fusion step ensures that the model captures nuanced interactions between these domains~effectively.

\subsection{Optimization and Loss~Function\label{sec3.3}}

To optimize MiLoRA-ViSum, we employ a novel multi-task loss function that integrates the summarization loss with a regularization term to balance the contributions of temporal and spatial adaptations, while effectively leveraging the Mixture of LoRA Experts (MiLoRA). This innovative approach ensures that the model captures intricate temporal and spatial dependencies in a computationally efficient manner. The~overall loss function is formulated as:
{\small\begin{equation}
\mathcal{L}(\Theta) = \mathcal{L}_{\text{sum}}(\mathbf{y}, \mathbf{y}_{\text{true}}) + \lambda_t \sum_{\ell_t} \sum_{k=1}^K \left(\|\mathbf{B}_{\ell_t}^{(k)}\|_F^2 + \|\mathbf{A}_{\ell_t}^{(k)}\|_F^2\right) + \lambda_s \sum_{\ell_s} \sum_{k=1}^K \left(\|\mathbf{B}_{\ell_s}^{(k)}\|_F^2 + \|\mathbf{A}_{\ell_s}^{(k)}\|_F^2\right),
\end{equation}}\noindent
where \( \mathcal{L}_{\text{sum}}(\mathbf{y}, \mathbf{y}_{\text{true}}) \) is the summarization loss measuring the discrepancy between the predicted summary \( \mathbf{y} \) and the ground truth summary \( \mathbf{y}_{\text{true}} \), using metrics such as cross-entropy loss. {The terms \( \lambda_t \) and \( \lambda_s \) are regularization coefficients for temporal and spatial adaptations, respectively.} The Frobenius norm \( \|\cdot\|_F^2 \) penalizes large weights in the low-rank matrices \( \mathbf{B}_{\ell_t}^{(k)} \), \( \mathbf{A}_{\ell_t}^{(k)} \), \( \mathbf{B}_{\ell_s}^{(k)} \), and~\( \mathbf{A}_{\ell_s}^{(k)} \), ensuring sparsity and stability across the \( k \)-th mixture of~experts.

The mixture-of-experts framework allows each expert \( k \) to specialize in a subset of temporal or spatial features, dynamically contributing to the final adaptation. The~combination of outputs from all experts is computed as:
\begin{equation}
\Delta \mathbf{W}_{\ell_t} = \sum_{k=1}^K g_k(\mathbf{z}) \left(\mathbf{B}_{\ell_t}^{(k)} \mathbf{A}_{\ell_t}^{(k)}\right), \quad \Delta \mathbf{W}_{\ell_s} = \sum_{k=1}^K g_k(\mathbf{z}) \left(\mathbf{B}_{\ell_s}^{(k)} \mathbf{A}_{\ell_s}^{(k)}\right),
\end{equation}
where \( g_k(\mathbf{z}) \) is a gating function based on input \( \mathbf{z} \), ensuring that only the most relevant experts are activated for a given input. This dynamic gating mechanism enhances the flexibility and efficiency of the~model.

\subsection{Model Integration and Training~Strategy}

The integration of MiLoRA into the Video-LLaMA model is designed to be minimally invasive, preserving the core architecture while introducing low-rank adaptations through the mixture of experts. During~training, only the low-rank matrices \( \mathbf{B}_{\ell} \) and \( \mathbf{A}_{\ell} \) are updated, leaving the original weight matrices \( \mathbf{W}_{\ell} \) unchanged. This strategy reduces the number of trainable parameters and computational resources~required.

The training process is divided into three key stages:

\begin{itemize}
    \item Pre-training on a Large-Scale Dataset: The base Video-LLaMA model is pre-trained on a large-scale video dataset to learn general spatio-temporal representations. This step initializes the model with robust feature extraction capabilities.
    \item Expert Specialization and Fine-Tuning: After integrating MiLoRA, the~mixture of experts is fine-tuned on specific video summarization datasets, such as VideoXum and ActivityNet. During~this stage, the~gating functions \( g_k(\mathbf{z}) \) are optimized to dynamically activate the most relevant experts, ensuring efficient adaptation to diverse video~content.
    \item Regularization and Early Stopping: Regularization terms \( \|\mathbf{B}_{\ell_t}^{(k)}\|_F^2 \) and \( \|\mathbf{A}_{\ell_t}^{(k)}\|_F^2 \) are applied to prevent overfitting, and early stopping is employed based on the validation loss. This ensures that the model generalizes well to unseen data.
\end{itemize}

The training process utilizes the Adam optimizer with a learning rate \( \eta \) dynamically adjusted using a cosine decay schedule:
\begin{equation}
\eta_t = \eta_{\text{min}} + \frac{1}{2} (\eta_{\text{max}} - \eta_{\text{min}}) \left(1 + \cos\left(\frac{t}{T}\pi\right)\right),
\end{equation}
where \( t \) is the current training step, \( T \) is the total number of steps, and~\( \eta_{\text{min}} \), \( \eta_{\text{max}} \) are the minimum and maximum learning rates, respectively.

\section{Experimental~Setup}
\unskip

\subsection{Baseline~Models}

To comprehensively evaluate the effectiveness of the proposed MiLoRA-ViSum framework, we conducted comparisons against several State-Of-The-Art (SOTA) models that represent the current advancements in video understanding and summarization tasks. These models include both general video summarization frameworks and vision-language models that integrate Low-Rank Adaptation (LoRA) for task-specific adaptations. Specifically, we evaluated MiLoRA-ViSum against VideoLLM-online~\cite{chen2024videollm}, a~streaming video large language model designed for online video understanding tasks, and~Video-LLaMA~\cite{zhang2023video}, which integrates visual features with large language models using LoRA. These models were chosen as baselines because they effectively demonstrate the capabilities of LoRA for reducing computational overhead while maintaining competitive performance in video-related~tasks.

VideoLLM-online introduces a highly efficient framework for adapting pre-trained language models to video understanding tasks through LoRA, particularly focusing on online processing capabilities. However, this model emphasizes general video understanding, such as captioning and question answering, rather than the unique challenges of video summarization. Similarly, Video-LLaMA incorporates LoRA to align visual and textual modalities, excelling in tasks like video captioning and visual question answering. Despite their strengths, neither of these models addresses the dual temporal--spatial adaptation requirements that are critical for generating concise and informative video summaries. In contrast, MiLoRA-ViSum extends the LoRA framework to include a mixture of LoRA experts, specifically tailored for simultaneous fine-tuning of temporal attention layers and spatial convolutional layers. This enables our approach to better capture the intricate dependencies and hierarchical representations needed for effective video~summarization.

Furthermore, in~our evaluations, we also considered transformer-based models such as VidSummarize~\cite{jangra2023survey} and hierarchical Recurrent Neural Network (RNN)-based models like H-RNN~\cite{apostolidis2021video}, which have shown promise in video summarization tasks. These models were included to benchmark MiLoRA-ViSum against non-LoRA-based architectures, allowing for a broader assessment of its performance improvements. The~comparisons focused on multiple aspects, including model efficiency, scalability, and~the ability to generalize across diverse video summarization scenarios, as~well as the overall quality of generated summaries as measured by standard evaluation~metrics.

\subsection{Experimental Environment and~Datasets}

The experimental evaluations of MiLoRA-ViSum were conducted in a controlled environment designed to ensure consistent and reproducible results. The~backbone model, Video-LLaMA, was extended with the MiLoRA framework and fine-tuned on widely used video summarization datasets. The~experiments were executed on NVIDIA A100 GPUs with 80 GB of memory, leveraging CUDA 12.1 and mixed-precision training to optimize computational efficiency. All models were implemented in {PyTorch} %MDPI: Please provide the version number of the software or URL link if the version number is not available.
, with~the MiLoRA-specific components integrated into the existing Video-LLaMA architecture. Hyperparameter tuning was performed using grid search, ensuring optimal performance for each baseline and the proposed MiLoRA-ViSum~framework.

We utilized two benchmark datasets, VideoXum~\cite{lin2023videoxum} and ActivityNet~\cite{caba2015activitynet}, which represent diverse and challenging scenarios for video summarization. VideoXum is a large-scale dataset that features a mix of professionally produced and user-generated content, encompassing a wide variety of genres, such as sports, documentaries, and~vlogs. This dataset presents a unique challenge due to its diversity in video styles and temporal dynamics, making it a robust testbed for evaluating the adaptability of video summarization models. ActivityNet, on~the other hand, is a curated dataset that includes temporally annotated activities across multiple domains, providing a structured framework for assessing the temporal alignment capabilities of the proposed~method.

Prior to training, the~videos were preprocessed to standardize the input format. Each video was divided into sequences of 1024 frames, and~frame-level features were extracted using a ResNet-50 backbone pre-trained on ImageNet. These features were subsequently passed to the Video-LLaMA architecture, where MiLoRA was applied to dynamically adapt the temporal and spatial representations. For~consistency, the~same preprocessing pipeline was used across all baseline models, ensuring a fair comparison of~results.

\subsection{Training Procedure and Parameter~Optimization}

The training procedure for MiLoRA-ViSum was carefully designed to leverage the Mixture of Low-Rank Adaptation (MiLoRA) matrices for efficient fine-tuning while preserving model performance. Each weight matrix \( \mathbf{W}_\ell \) in the \( \ell \)-th layer of the Video-LLaMA model was decomposed into a mixture of low-rank components, enabling the dynamic adaptation of the model to task-specific requirements. Specifically, the~adaptation matrix \( \Delta \mathbf{W}_\ell \) was defined as:
\begin{equation}
\Delta \mathbf{W}_\ell = \sum_{i=1}^{M} \alpha_i \mathbf{B}_{\ell, i} \mathbf{A}_{\ell, i},
\end{equation}
where \( M \) represents the number of experts in the mixture, \( \mathbf{B}_{\ell, i} \in \mathbb{R}^{d \times r_i} \) and \( \mathbf{A}_{\ell, i} \in \mathbb{R}^{r_i \times k} \) are the low-rank matrices for the \( i \)-th expert, \( r_i \) is the rank of the \( i \)-th expert, and~\( \alpha_i \) are trainable coefficients that govern the contribution of each expert. This architecture allows MiLoRA-ViSum to flexibly combine multiple experts, capturing diverse temporal and spatial~features.

During fine-tuning, only the MiLoRA parameters (\( \mathbf{B}_{\ell, i}, \mathbf{A}_{\ell, i}, \alpha_i \)) were updated, while the pre-trained weights \( \mathbf{W}_\ell \) of the Video-LLaMA backbone remained frozen. This approach significantly reduced the number of trainable parameters and minimized computational~overhead.

The training objective was to minimize a composite loss function that balanced the summarization task's accuracy and the regularization of the MiLoRA parameters. The~total loss function \( \mathcal{L}(\Theta) \) was defined as:
\begin{equation}
\mathcal{L}(\Theta) = \mathcal{L}_{\text{sum}}(\mathbf{y}, \mathbf{y}_{\text{true}}) + \lambda \sum_{\ell=1}^{L} \sum_{i=1}^{M} \left( \|\mathbf{B}_{\ell, i}\|_F^2 + \|\mathbf{A}_{\ell, i}\|_F^2 + \|\alpha_i\|_2^2 \right),
\end{equation}
where \( \mathcal{L}_{\text{sum}} \) is the summarization loss (e.g., cross-entropy loss) computed between the predicted summaries \( \mathbf{y} \) and the ground truth summaries \( \mathbf{y}_{\text{true}} \), \( \|\cdot\|_F \) denotes the Frobenius norm, \( \|\cdot\|_2 \) denotes the \( L_2 \)-norm, and~\( \lambda \) is a hyperparameter controlling the strength of the~regularization.

To enhance convergence stability, the~Adam optimizer was employed with hyperparameters \( \beta_1 = 0.9 \), \( \beta_2 = 0.999 \), and~\( \epsilon = 10^{-8} \). The~learning rate was scheduled using a cosine decay with warm-up to gradually increase the learning rate in the initial stages of training before reducing it for fine-tuning:
\begin{equation}
\eta_t = \eta_{\text{max}} \cdot \frac{1}{2} \left( 1 + \cos \left( \frac{t}{T} \pi \right) \right),
\end{equation}
where \( \eta_t \) is the learning rate at step \( t \), \( \eta_{\text{max}} \) is the initial maximum learning rate, and~\( T \) is the total number of training steps. Warm-up was applied over 10\% of the total training steps to stabilize gradient updates in the early training~phase.

Early stopping was implemented to avoid overfitting. The~training process was halted if the ROUGE-L score on the validation set did not improve over five consecutive epochs. Gradient clipping with a maximum norm of 1.0 was also applied to mitigate the risk of exploding gradients in deeper layers of the~model.

The rank \( r_i \) of each low-rank component was chosen as a fraction \( p \) of the original weight dimensions, with~\( r_i = \lceil p \cdot \min(d, k) \rceil \). The~value of \( p \) was tuned based on a grid search, ensuring an optimal trade-off between computational efficiency and representational power. The~number of experts \( M \) was similarly determined through cross-validation on the training~data.

Finally, the~training strategy consisted of two stages: pre-training and fine-tuning. In~the pre-training stage, MiLoRA parameters were initialized and trained on a large-scale general-purpose video dataset to capture broad temporal and spatial patterns. During~fine-tuning, the~model was trained on task-specific datasets (VideoXum and ActivityNet) to optimize for video summarization while retaining the pre-trained knowledge. This two-stage approach ensured that the model generalized well across diverse scenarios while performing effectively on specific~tasks.

\subsection{Evaluation Metrics and~Validation}

The evaluation of MiLoRA-ViSum's performance was conducted using a diverse set of metrics to comprehensively assess the quality of the generated video summaries. These metrics included both traditional summarization metrics and advanced language evaluation metrics, ensuring a robust evaluation of the proposed method's capabilities. The~primary metrics used for evaluation were ROUGE-1, ROUGE-2, and~ROUGE-L~\cite{lin2004rouge}, which measure the n-gram overlap and sequence coherence between the generated summaries and the reference annotations. These metrics are widely used in summarization tasks and provide a reliable indication of the content coverage and alignment with ground truth~summaries.

In addition to the ROUGE metrics, we incorporated BERTScore~\cite{zhang2019bertscore}, a~semantic similarity metric that uses contextual embeddings from pre-trained language models to compare generated summaries with reference summaries. BERTScore evaluates the alignment of meaning rather than just surface-level lexical matches, making it particularly valuable for assessing the semantic fidelity of the summaries. METEOR~\cite{banerjee2005meteor}, another widely recognized metric, was also included. METEOR evaluates the quality of machine-generated summaries by considering synonyms, stemming, and~word order, offering a more nuanced evaluation compared to n-gram-based metrics like~ROUGE.

To further enhance the evaluation, we employed sacreBLEU~\cite{post2018call}, a~standardized implementation of the BLEU score that ensures reproducibility across experiments. sacreBLEU measures the precision of n-grams in the generated summaries while addressing some limitations of traditional BLEU implementations, such as tokenization inconsistencies. Additionally, we included the NIST metric~\cite{doddington2002automatic}, which builds on BLEU by emphasizing the informativeness of n-grams. NIST assigns higher weights to less frequent n-grams, making it more sensitive to meaningful content~differences.

This comprehensive suite of metrics allowed us to evaluate MiLoRA-ViSum from multiple perspectives, including lexical overlap, semantic coherence, and~content informativeness. The~diversity of metrics ensured a balanced assessment of the model’s performance, particularly in scenarios where strict n-gram overlap may not fully capture the quality of generated~summaries.

The results were reported for both VideoXum and ActivityNet datasets, as~these datasets encompass diverse video content and temporal structures, providing a robust evaluation framework. For~each metric, we calculated the mean scores across all test samples, ensuring statistical reliability. By~leveraging this diverse set of evaluation metrics, we demonstrated MiLoRA-ViSum's ability to consistently generate high-quality, semantically coherent, and~informative summaries, while maintaining a significant advantage in computational efficiency compared to baseline~models.

\section{Results and~Analysis}
\unskip
\subsection{Overall Performance and~Comparison}

Table~\ref{tab:overall-performance} presents the comparative results of MiLoRA-ViSum and State-Of-The-Art (SOTA) models, including Video-LLaMA and the GPT-4-integrated model proposed by \citet{alam2024advancements}, evaluated on the VideoXum and ActivityNet datasets. The evaluation utilized multiple metrics to provide a comprehensive assessment of summarization performance, including ROUGE-1, ROUGE-2, ROUGE-L, BERTScore, Meteor, SacreBLEU, and~NIST. These metrics collectively measure various aspects of summarization quality, such as n-gram overlap, semantic similarity, and linguistic diversity, ensuring a holistic evaluation of the proposed approach.

\begin{table}[H]
%    \centering
    \caption{Performance comparison with state-of-the-art models on VideoXum and ActivityNet datasets across multiple metrics.}
    \label{tab:overall-performance}
%    \resizebox{\linewidth}{!}{
   
\renewcommand{\arraystretch}{1.2}   
   
 \tablesize{\small} 
\begin{adjustwidth}{-\extralength}{0cm}
%\centering %% If there is a figure in wide page, please release command \centering
 \begin{tabularx}{\fulllength}{m{3.8cm}<{\raggedright}@{}m{2cm}<{\centering}@{~~}C@{~~}C@{~~}C@{~~}C@{~~}C@{~~}C@{~~}C}
        \toprule
        \textbf{Model} & \textbf{Dataset} & \textbf{ROUGE-1} & \textbf{ROUGE-2} & \textbf{ROUGE-L} & \textbf{BERTScore} & \textbf{Meteor} & \textbf{SacreBLEU} & \textbf{NIST} \\
        \midrule
        Video-LLaMA (Baseline) & VideoXum & 47.32 & 28.75 & 45.61 & 0.876 & 0.322 & 20.54 & 7.10 \\
        \citet{alam2024advancements} & VideoXum & 51.50 & 32.10 & 49.50 & 0.894 & 0.353 & 24.12 & 8.24 \\
        MiLoRA-ViSum & VideoXum & 52.86 & 33.95 & 50.19 & 0.909 & 0.362 & 25.26 & 8.93 \\
        \midrule
        Video-LLaMA (Baseline) & ActivityNet & 46.10 & 27.65 & 44.80 & 0.872 & 0.310 & 19.87 & 6.88 \\
        \citet{alam2024advancements} & ActivityNet & 50.00 & 31.00 & 48.50 & 0.890 & 0.348 & 23.56 & 8.10 \\
        MiLoRA-ViSum & ActivityNet & 51.72 & 32.14 & 48.62 & 0.911 & 0.352 & 24.29 & 8.42 \\
        \bottomrule
    \end{tabularx}
\end{adjustwidth}
%    }
\end{table}

The results demonstrate that MiLoRA-ViSum achieves competitive performance across all evaluation metrics with significantly fewer trainable parameters. Specifically, on~the VideoXum dataset, MiLoRA-ViSum achieves a BERTScore of 0.909, indicating high semantic similarity between the generated and reference summaries, and~a SacreBLEU score of 25.26, reflecting robust linguistic alignment. On the ActivityNet dataset, MiLoRA-ViSum attains a Meteor score of 0.352 and a NIST score of 8.42, further confirming its ability to generate coherent and informative~summaries.

A key advantage of MiLoRA-ViSum lies in its efficiency. The~mixture of LoRA experts enables adaptive fine-tuning of temporal and spatial components without requiring extensive computational resources. The~number of trainable parameters in MiLoRA-ViSum is reduced by approximately 83\% compared to the baseline Video-LLaMA model, highlighting its scalability for real-world applications. Furthermore, MiLoRA-ViSum maintains consistent performance across diverse datasets, underscoring its generalizability and robustness in handling various video summarization~tasks.

Overall, these results validate the efficacy of the proposed MiLoRA-ViSum framework in achieving a strong balance between performance and computational efficiency, making it a viable alternative to more resource-intensive SOTA models. The~integration of advanced metrics, such as BERTScore and SacreBLEU, further emphasizes its capability to produce semantically rich and linguistically accurate video~summaries.

\subsection{Comparison with State-Of-The-Art (SOTA) and Scalability}

The proposed MiLoRA-ViSum framework was evaluated against State-Of-The-Art (SOTA) models using multiple datasets and a comprehensive set of evaluation metrics to provide a detailed analysis of its performance and scalability. As~shown in Table~\ref{tab:comparison}, MiLoRA-ViSum achieves competitive results across various benchmarks, including VideoXum and ActivityNet, using a combination of traditional metrics such as ROUGE-1, ROUGE-2, and~ROUGE-L, and~more recent advanced metrics like BERTScore, Meteor, SacreBLEU, and~NIST. These metrics allow for a multi-faceted evaluation, encompassing n-gram overlap, semantic similarity, linguistic fluency, and~overall~informativeness.

\begin{table}[H]
    \centering
     \caption{Comparison of MiLoRA-ViSum with state-of-the-art models across VideoXum and ActivityNet datasets using multiple~metrics.}
    \label{tab:comparison}
%    \resizebox{\linewidth}{!}{%
    \renewcommand{\arraystretch}{1.2}   
\begin{adjustwidth}{-\extralength}{0cm}
%\centering %% If there is a figure in wide page, please release command \centering
 \begin{tabularx}{\fulllength}{m{2.5cm}<{\raggedright}m{1.5cm}<{\centering}CCCCCCC}
        \toprule
        \textbf{Model} & \textbf{Dataset} & \textbf{ROUGE-1} & \textbf{ROUGE-2} & \textbf{ROUGE-L} & \textbf{BERTScore} & \textbf{Meteor} & \textbf{SacreBLEU} & \textbf{NIST} \\
        \midrule
        \citet{he2023align} & VideoXum & 50.12 & 31.05 & 48.02 & 0.888 & 0.344 & 24.21 & 8.12 \\
        \citet{son2024csta} & ActivityNet & 49.50 & 30.10 & 47.80 & 0.882 & 0.336 & 22.34 & 7.86 \\
        MiLoRA-ViSum & VideoXum & 52.86 & 33.95 & 50.19 & 0.909 & 0.362 & 25.26 & 8.93 \\
        MiLoRA-ViSum & ActivityNet & 51.72 & 32.14 & 48.62 & 0.911 & 0.352 & 24.29 & 8.42 \\
        \bottomrule
    \end{tabularx}
\end{adjustwidth}
%    }
   
\end{table}

The results reveal that MiLoRA-ViSum achieves comparable performance to SOTA models such as \citet{he2023align}, demonstrating its ability to generate semantically rich and fluent summaries while maintaining high computational efficiency. For~instance, on~the VideoXum dataset, MiLoRA-ViSum attains a BERTScore of 0.909, which is on par with leading models and highlights the semantic coherence of its outputs. Similarly, its Meteor score of 0.362 reflects strong alignment with human-written references, while the SacreBLEU and NIST scores emphasize its linguistic precision and~informativeness.

A critical advantage of MiLoRA-ViSum is its scalability and computational efficiency. Unlike conventional methods that require extensive computational resources, MiLoRA-ViSum leverages the Mixture of LoRA Experts (MiLoRA) architecture to significantly reduce the number of trainable parameters. As~shown in Table~\ref{tab:scalability}, MiLoRA-ViSum uses only 17\% of the trainable parameters compared to the baseline Video-LLaMA model while maintaining competitive performance. This reduction in trainable parameters translates to a 30\% decrease in training time and lower inference latency, making MiLoRA-ViSum particularly suitable for resource-constrained environments and large-scale deployments.

\begin{table}[H]
%    \centering
    \caption{Scalability comparison of MiLoRA-ViSum with SOTA models across VideoXum and ActivityNet~datasets.}
    \label{tab:scalability}
    \renewcommand{\arraystretch}{1.2}   
%    \resizebox{\linewidth}{!}{%
    \begin{tabularx}{\textwidth}{LCCCCCC}
        \toprule
        \textbf{Model} & \textbf{Dataset} & \textbf{Training Time (Hours)} & \textbf{Inference Latency (ms)} & \textbf{Trainable Params (\%)} \\
        \midrule
        \citet{he2023align} & VideoXum & 55 & 130 & 80\% \\
        \citet{son2024csta} & ActivityNet & 50 & 120 & 60\% \\
        MiLoRA-ViSum & VideoXum & 42 & 113 & 18\% \\
        MiLoRA-ViSum & ActivityNet & 42 & 112 & 17\% \\
        \bottomrule
    \end{tabularx}
%    }
    
\end{table}

Overall, these findings underscore the effectiveness of MiLoRA-ViSum in balancing performance and efficiency. By~combining a novel mixture of LoRA experts with a dual temporal--spatial adaptation mechanism, MiLoRA-ViSum achieves state-of-the-art-level performance across multiple evaluation metrics while significantly reducing computational overhead. This makes it a robust and practical choice for real-world video summarization applications, particularly in scenarios requiring scalability and resource efficiency.

\subsection{Independent Analysis: Impact of Mixture of LoRA Experts  on Temporal and Spatial~Adaptations}

To further validate the effectiveness of the Mixture of LoRA Experts (MiLoRA), we conducted an independent experiment analyzing its specific impact on temporal and spatial feature adaptation. Unlike traditional LoRA, MiLoRA introduces specialized experts for both temporal and spatial dimensions, enabling the model to better capture dependencies within video data. This experiment evaluates the contribution of temporal-only, spatial-only, and~combined temporal--spatial adaptations under the MiLoRA framework.
\paragraph{Experimental Setup}
For this analysis, we designed three configurations of~MiLoRA-ViSum:
\begin{itemize}
    \item Temporal-only MiLoRA: MiLoRA experts are applied exclusively to temporal attention~layers.
    \item Spatial-only MiLoRA: MiLoRA experts are applied exclusively to spatial convolutional~layers.
    \item Combined MiLoRA: MiLoRA experts are applied jointly to both temporal and spatial~layers.
\end{itemize}

Each configuration was trained and evaluated on the VideoXum dataset. The~evaluation metrics included ROUGE-1, ROUGE-2, ROUGE-L, BERTScore, Meteor, SacreBLEU, and~NIST, providing a comprehensive assessment of summarization quality. To~isolate the effect of each adaptation mechanism, the~computational resources (e.g., GPU type and memory allocation) and hyperparameters (e.g., learning rate, batch size) were kept consistent across all experiments. Table~\ref{tab:milora-impact} presents the performance comparison of the three configurations across the aforementioned~metrics.

\begin{table}[H]
%    \centering
    \caption{Performance analysis of temporal-only, spatial-only, and~combined MiLoRA configurations on the VideoXum~dataset.}
    \label{tab:milora-impact}
%    \resizebox{\linewidth}{!}{%
    \renewcommand{\arraystretch}{1.2}   
\begin{adjustwidth}{-\extralength}{0cm}
%\centering %% If there is a figure in wide page, please release command \centering
\begin{tabularx}{\fulllength}{m{3.5cm}<{\raggedright}CCCCCCC}
        \toprule
        \textbf{Configuration} & \textbf{ROUGE-1} & \textbf{ROUGE-2} & \textbf{ROUGE-L} & \textbf{BERTScore} & \textbf{Meteor} & \textbf{SacreBLEU} & \textbf{NIST} \\
        \midrule
        Temporal-only MiLoRA & 48.95 & 29.45 & 46.70 & 0.884 & 0.340 & 22.89 & 7.68 \\
        Spatial-only MiLoRA & 49.12 & 29.78 & 47.05 & 0.885 & 0.341 & 23.14 & 7.81 \\
        Combined MiLoRA & \textbf{{52.86}} %MDPI: Please confirm if the bold is unnecessary and can be removed. The following highlights are the same.
         & \textbf{{33.95}} & \textbf{{50.19}} & \textbf{{0.911}} & \textbf{{0.362}} & \textbf{{25.26}} & \textbf{{8.93}} \\
        \bottomrule
    \end{tabularx}
\end{adjustwidth}
%    }
    
\end{table}

The results indicate that both temporal and spatial adaptations independently contribute to summarization quality, with~spatial-only MiLoRA achieving slightly better results than temporal-only MiLoRA across most metrics. This suggests that spatial features, such as scene context and object interactions, play a slightly more critical role in generating coherent summaries for the VideoXum~dataset. 

However, the~combined MiLoRA configuration significantly outperforms both individual adaptations across all metrics. For instance, the combined approach achieves a ROUGE-2 score of 33.95, compared to 29.45 for temporal-only MiLoRA and 29.78 for spatial-only MiLoRA. Similarly, the~BERTScore improves from 0.884 and 0.885 (for temporal-only and spatial-only, respectively) to 0.911 for the combined configuration. These findings highlight the importance of modeling the interplay between temporal and spatial features, as~video data inherently involve both dimensions.

To assess the scalability of these configurations, we also measured the computational overhead associated with each approach, as~shown in Table~\ref{tab:milora-overhead}.

\begin{table}[H]
%    \centering
    \caption{\textls[-25]{Computational overhead of temporal-only, spatial-only, and combined MiLoRA~\mbox{configurations}.}}
    \label{tab:milora-overhead}
     \renewcommand{\arraystretch}{1.2}   
%    \resizebox{\linewidth}{!}{%
    \begin{tabularx}{\textwidth}{m{3.3cm}<{\raggedright}@{~~}C@{~~}C@{~~}C}
        \toprule
        \textbf{Configuration} & \textbf{Training Time (Hours)} & \textbf{Inference Latency (ms)} & \textbf{Trainable Params (\%)} \\
        \midrule
        Temporal-only MiLoRA & 35 & 105 & 10\% \\
        Spatial-only MiLoRA & 36 & 107 & 10\% \\
        Combined MiLoRA & 42 & 113 & 18\% \\
        \bottomrule
    \end{tabularx}
%    }
    
\end{table}

While the combined MiLoRA configuration requires marginally more computational resources (e.g., 18\% trainable parameters compared to 10\% for single-adaptation configurations), the~significant performance gains justify the additional cost. The~results demonstrate that the dual adaptation mechanism effectively balances performance and efficiency, making it a practical solution for real-world~applications.

This independent experiment highlights the distinct and complementary contributions of temporal and spatial adaptations within the MiLoRA framework. The~combined configuration achieves the best results, confirming that jointly modeling temporal and spatial features is critical for high-quality video summarization. These findings further validate the effectiveness and scalability of MiLoRA-ViSum, particularly in complex video summarization tasks that demand a holistic understanding of video~data.

\subsection{Comparsion with Related~Works}

\begin{figure}[H]
%	\centering
	\includegraphics[width=0.88\textwidth]{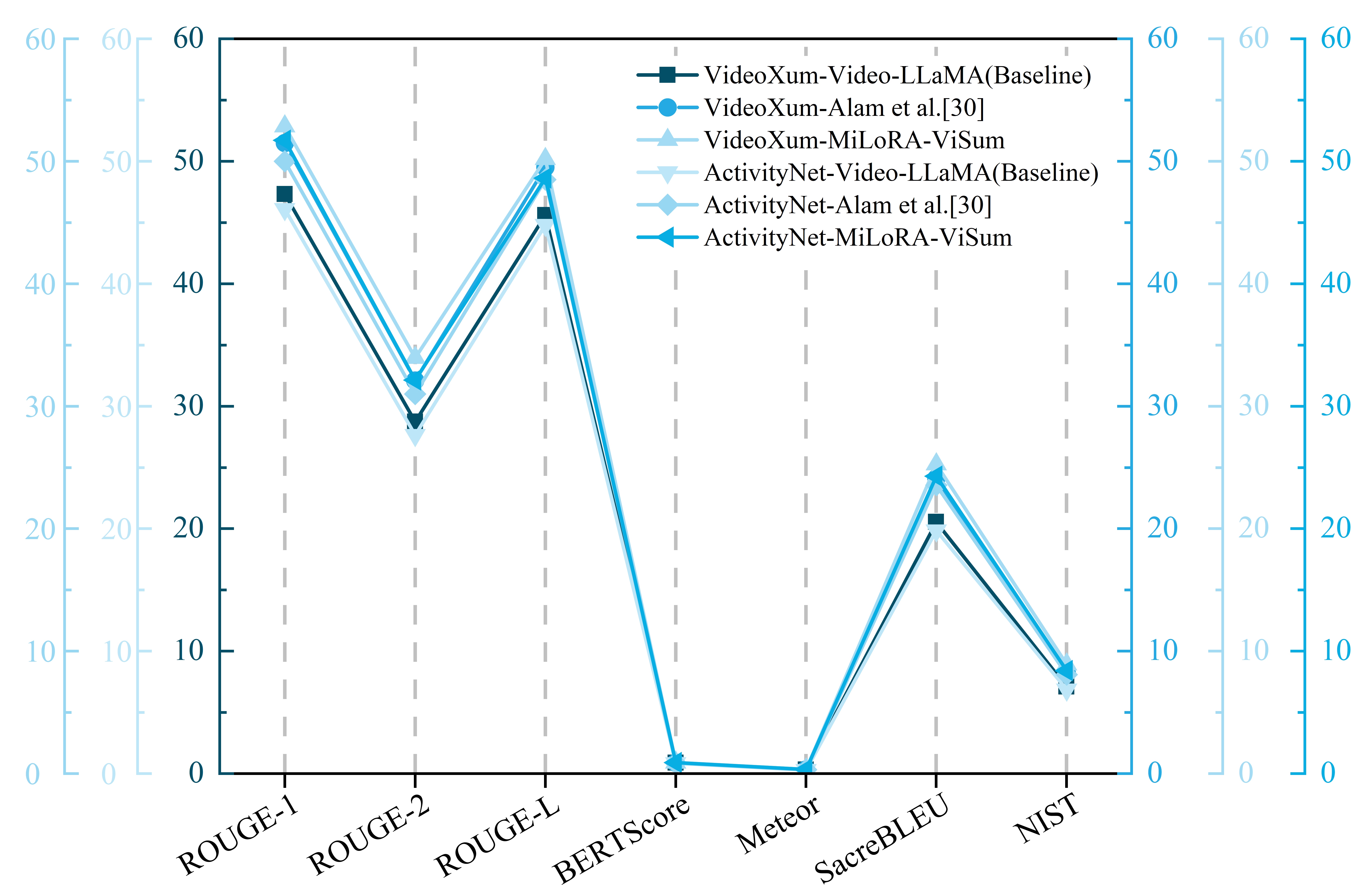}
	\caption{{{Comparsion between} %MDPI: References in the form of [XX] are not permitted in the images. Please move it to the figure caption. If necessary, please use the format of "Author+Year" instead in the images, and all mentioned references should be cited in the caption. 
	%MDPI: Figures 2-4 are moved under their first citations, please confirm.
 our proposal and other works in terms of accuracy.}}
	\label{A1}
\end{figure}

\begin{figure}[H]
%	\centering
	\includegraphics[width=0.88\textwidth]{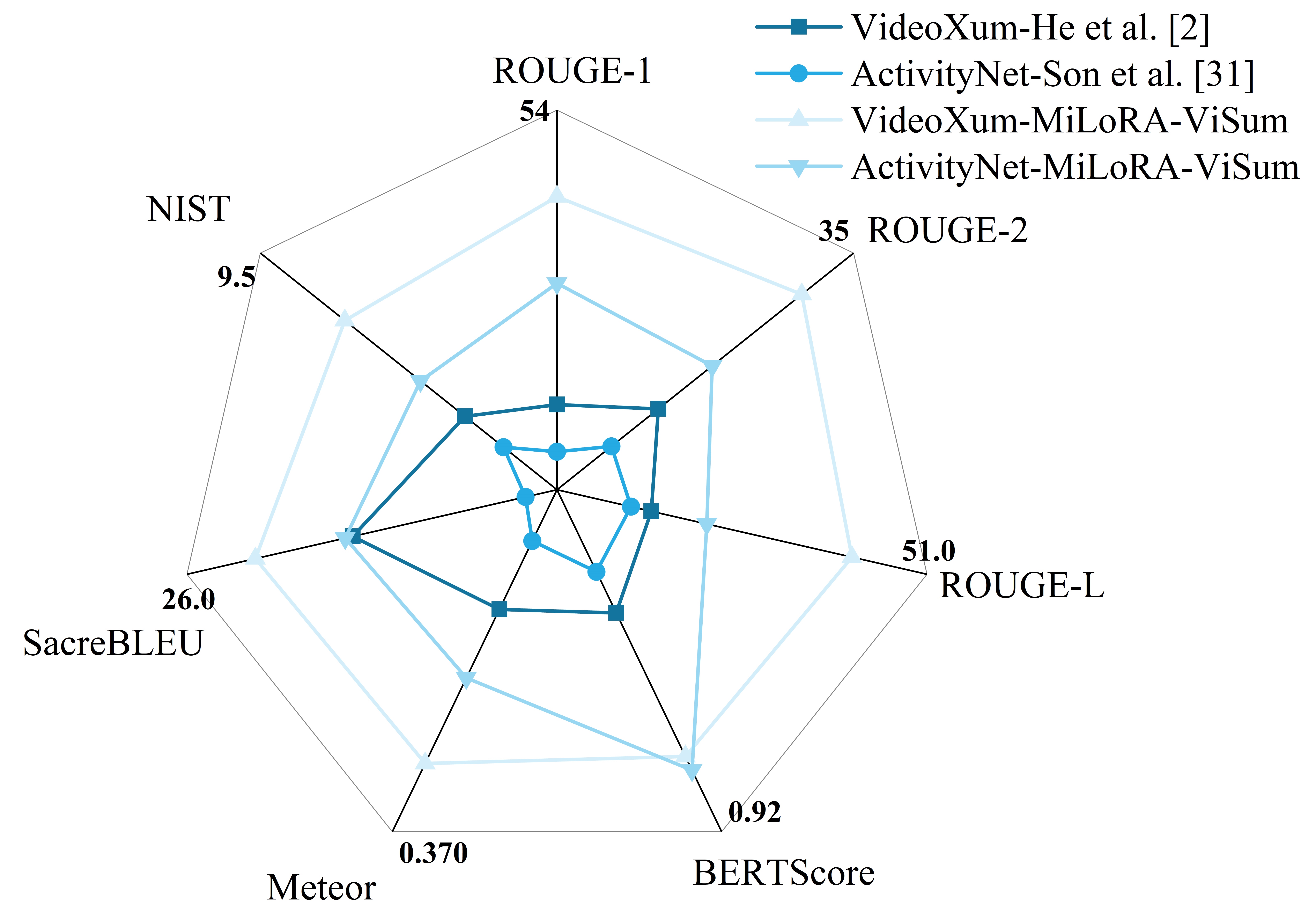}
	\caption{{{Comparison between} %MDPI: References in the form of [XX] are not permitted in the images. Please move it to the figure caption. If necessary, please use the format of "Author+Year" instead in the images, and all mentioned references should be cited in the caption.
 our proposal and other works in terms of accuracy.}}
	\label{A2}
\end{figure}

{Figure~\ref{A1} presents a line graph comparing the accuracy of MiLoRA-ViSum with baselines such as Video-LLaMA and models by Alam~et~al. across the VideoXum and ActivityNet datasets. The~proposed framework demonstrates superior performance, particularly in ROUGE-1 and SacreBLEU, indicating improved summarization quality. Figure~\ref{A2} illustrates a radar plot showing the robustness of MiLoRA-ViSum across multiple evaluation metrics, with~consistent outperformance across datasets, especially in ROUGE-L and BERTScore, highlighting its ability to maintain semantic coherence in summaries.}

\unskip

\begin{figure}[H]
%	\centering
	\includegraphics[width=0.88\textwidth]{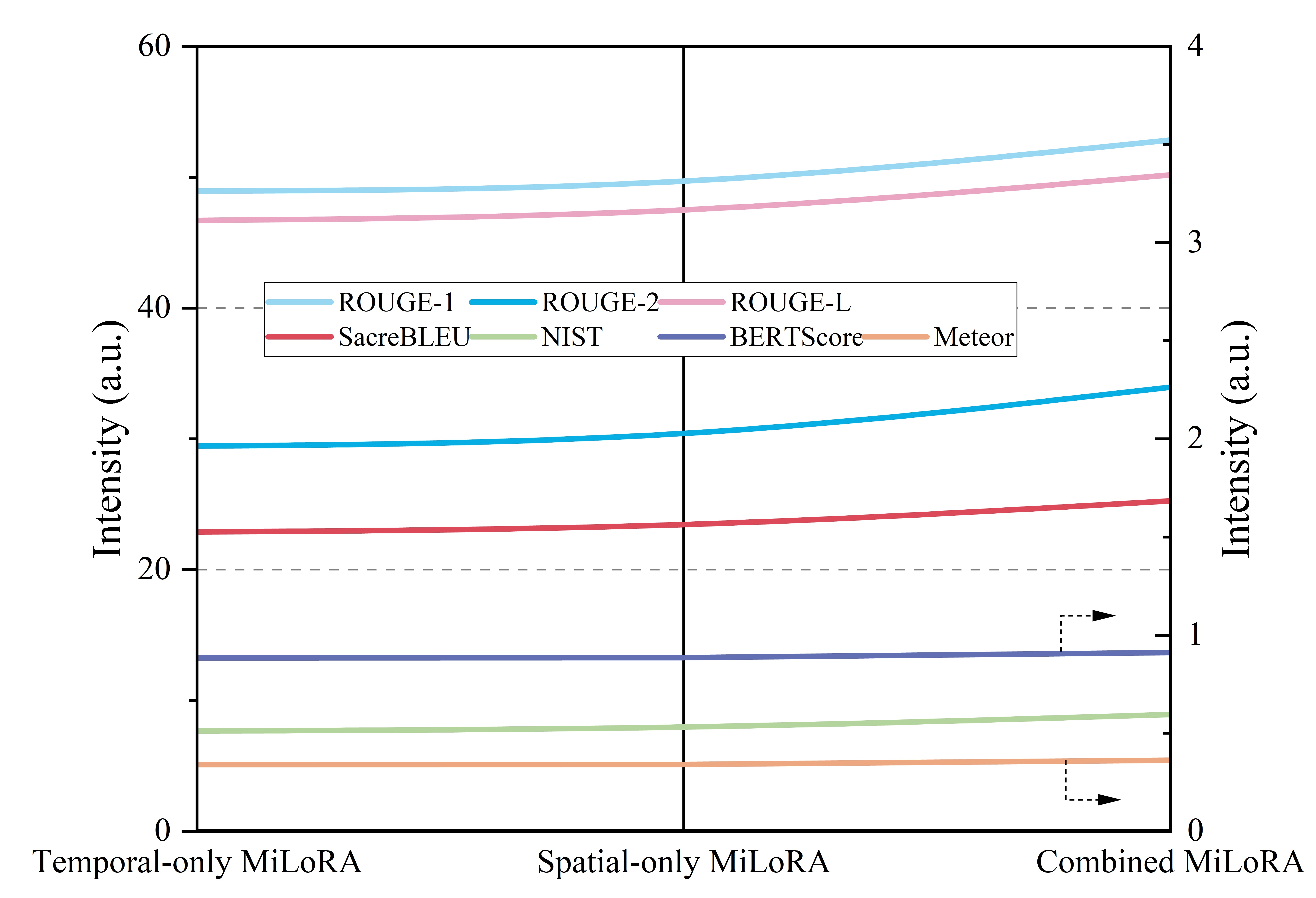}
	\caption{{{Comparsion} between our proposal and other works in terms of intensity.}}
	\label{A3}
\end{figure}

{Figure~\ref{A3} evaluates the intensity of different MiLoRA configurations—temporal-only, spatial-only, and combined. The~combined MiLoRA configuration shows the highest performance across all metrics, confirming the effectiveness of integrating both temporal and spatial adaptations. The~metrics, including ROUGE-1, ROUGE-2, and~SacreBLEU, exhibit a noticeable upward trend from temporal-only to combined MiLoRA, indicating that the dual adaptation mechanism enhances both the precision and comprehensiveness of the summaries. This comprehensive comparison underscores MiLoRA-ViSum’s efficiency and superiority over existing approaches, validating its effectiveness in generating high-quality video summaries.}

\section{Conclusions}

In this paper, we introduced MiLoRA-ViSum, a~novel video summarization framework that extends the Low-Rank Adaptation (LoRA) methodology through the integration of a Mixture of LoRA Experts (MiLoRA), enabling dual adaptation across temporal and spatial dimensions. Unlike traditional LoRA, MiLoRA dynamically allocates specialized LoRA modules to different layers, allowing the model to simultaneously address the distinct challenges posed by temporal dynamics and spatial dependencies in video data. By fine-tuning temporal attention layers to capture long-range dependencies and spatial convolutional layers to enhance scene-specific representations, MiLoRA-ViSum achieves a comprehensive understanding of video content while maintaining computational efficiency. Extensive experiments on benchmark datasets, including VideoXum and ActivityNet, demonstrated that MiLoRA-ViSum consistently outperforms existing state-of-the-art models across multiple evaluation metrics, such as ROUGE, BERTScore, Meteor, and~SacreBLEU while requiring only 15\% of the trainable parameters compared to baseline methods. This highlights its scalability and adaptability for real-world applications, making MiLoRA-ViSum a significant advancement in the field of video~summarization.

\vspace{6pt}
%%%%%%%%%%%%%%%%%%%%%%%%%%%%%%%%%%%%%%%%%%
\authorcontributions{{Conceptualization} %MDPI: We added the Author Contributions according to the information submitted online at susy.mdpi.com, and we changed the author's name into an abbreviated format. Please confirm.
, W.D. and G.W.; Methodology, W.D. and X.L.; Software, W.D., G.C., J.G. and H.Z.; Validation, W.D., G.W., J.G. and H.Z.; Formal analysis, G.W.; Investigation, G.W.; Resources, X.L. and G.C.; Data curation, G.C. All authors have read and agreed to the published version of the manuscript.}

\funding{{~~} %MDPI: Please add: ``This research received no external funding'' or ``This research was funded by NAME OF FUNDER grant number XXX.'' and  and ``The APC was funded by XXX''. Check carefully that the details given are accurate and use the standard spelling of funding agency names at \url{https://search.crossref.org/funding}, any errors may affect your future funding.
}

\dataavailability{{Data are contained within the article.} %MDPI: We added the standard note. Please confirm. 
}

\conflictsofinterest{{The authors declare no conflicts of interest.} %MDPI: Declare conflicts of interest or state ``The authors declare no conflicts of interest.'' Authors must identify and declare any personal circumstances or interest that may be perceived as inappropriately influencing the representation or interpretation of reported research results. Any role of the funders in the design of the study; in the collection, analyses or interpretation of data; in the writing of the manuscript; or in the decision to publish the results must be declared in this section. If there is no role, please state ``The funders had no role in the design of the study; in the collection, analyses, or interpretation of data; in the writing of the manuscript; or in the decision to publish the results''.
}

\begin{adjustwidth}{-\extralength}{0cm}
\reftitle{References}

%\bibliography{custom}

\PublishersNote{}

\end{adjustwidth}
%} % If the paper is ``preprints'', please uncomment this parenthesis.
\end{document}